\documentclass[12pt,3p]{elsarticle}
\usepackage{tabularx}
\usepackage{graphicx}
\usepackage{geometry}
\usepackage{pifont} 
\usepackage{array}
\usepackage{booktabs} 
\usepackage{ragged2e}
\usepackage{amsmath}
\usepackage{booktabs}
\usepackage{makecell}
\usepackage{url}
\usepackage{balance}
\usepackage{amsmath}
\usepackage{subfigure}
\usepackage{balance}
\usepackage{color}
\usepackage{comment}
\usepackage{array}
\usepackage{lipsum}
\usepackage{caption}
\usepackage{amsmath}
\usepackage{array}
\usepackage{pifont} 
\usepackage{booktabs} 

\usepackage{xcolor}     
\usepackage{multirow}   
\usepackage{booktabs}   
\usepackage{amssymb}    
\usepackage{caption}   
\usepackage{amsmath}   
\usepackage[table]{xcolor} 
\usepackage{colortbl}      
\usepackage{blindtext}
\usepackage{enumitem}
\usepackage{lipsum}
\usepackage{capt-of}
\usepackage{tikz}
\usetikzlibrary{arrows,positioning,automata}
\usepackage{balance}
\usepackage{amsmath}
\usepackage[T1]{fontenc}
\usepackage{algorithm}
\usepackage{algpseudocode}
\usepackage{subfigure}

\usepackage[utf8]{inputenc}
\usepackage{balance}
\usepackage{multirow}
\usepackage{multicol}
\usepackage{color}
\usepackage{comment}
\usepackage{array}
\usepackage{soul}
\usepackage[justification=centering]{caption}
\usepackage[lighttt]{lmodern}
\usepackage{libertine}
\usepackage{xcolor}
\usepackage[utf8]{inputenc}
\definecolor{turquoise}{RGB}{64, 224, 208}
\definecolor{darkblue}{RGB}{0, 0, 139}
\definecolor{green}{RGB}{0, 128, 0}
\usepackage{subcaption}
\usepackage{amssymb}
\usepackage{float}
\usepackage{hyperref}
\usepackage[perpage]{footmisc}
\usepackage{hyperref} 
\usepackage{zref-user}
\definecolor{bluee}{RGB}{45, 104, 196}
\definecolor{yellow-or}{RGB}{247, 167, 12}
\journal{}
\begin{document}
\begin{frontmatter}
\date{}
\title{Two Stage Context Learning with Large Language Models for Multimodal Stance Detection on Climate Change}
\date{ }
\author[label1]{Lata Pangtey}

\address[label1]{Department of Computer Science and Engineering, Indian Institute of Technology (IIT) Indore, Indore 453552, India}
\address[label2]{Chaitanya Bharathi Institute of Technology, Gandipet, Hyderabad, 500075}

\ead{ms2304101009@iiti.ac.in}

\author[label2]{Omkar Kabde}

\ead{omkarkabde@gmail.com }
\author[label1]{Shahid Shafi Dar}
\ead{phd2201201004@iiti.ac.in}
\author[label1]{Nagendra Kumar\corref{cor1}}
\ead{nagendra@iiti.ac.in}
\cortext[cor1]{Corresponding author}
\begin{abstract}
With the rapid proliferation of information across digital platforms, stance detection has emerged as a pivotal challenge in social media analysis. While most of the existing approaches focus solely on textual data, real-world social media content increasingly combines text with visual elements creating a need for advanced  multimodal methods. To address this gap, we propose a multimodal stance detection framework that integrates textual and visual information through a hierarchical fusion approach. Our method first employs a Large Language Model to retrieve stance-relevant summaries from source text, while a domain-aware image caption generator interprets visual content in the context of the target topic. These modalities are then jointly modeled along with the reply text, through a specialized transformer module that captures interactions between the texts and images. The proposed modality fusion framework integrates diverse modalities to facilitate robust stance classification.
We evaluate our approach on the MultiClimate dataset, a benchmark for climate change-related stance detection containing aligned video frames and transcripts. We achieve accuracy of 76.2\%, precision of 76.3\%, recall of 76.2\% and F1-score of 76.2\%, respectively, outperforming existing state-of-the-art approaches.
\end{abstract}

\end{frontmatter}
\section{Introduction}
Stance detection is the task of identifying an author's position (support, oppose, or neutral) concerning a specific theme or claim.
Stance detection is an important task in natural language processing and social media analysis, as it plays a key role in understanding public opinion, analyzing discourse, and combating misinformation~\cite{PANGTEY2025111109}.
Accurate stance detection plays a vital role in addressing critical societal issues such as climate change, public health, and the spread of misinformation. In these contexts, understanding public stance toward specific claims or topics enables researchers, policymakers, and social platforms to analyze public opinion trends and detect polarizing or controversial content. This is particularly important in the age of social media, where user-generated content often shapes public discourse and perception. 

Table~\ref{tab:stance_examples} presents three illustrative examples demonstrating how stance toward climate change is expressed through both textual and visual content. 
In Example 1, the supportive stance is evident in the topic title ``Climate Change: Mitigate or Adapt'' and the transcript promoting adaptation and mitigation strategies. The neutral component poses a question about local impacts, while the opposing stance acknowledges the inevitability of current climate effects despite mitigation efforts.
Example 2 shows support through a transcript that highlights NASA's scientific consensus on climate change. The neutral transcript introduces the speaker’s credentials, and the opposing stance stresses the global and long-term nature of the problem, implying limited individual agency.
{Example 3} includes a scientific observation in support, stating methane decreases and CO\textsubscript{2} increases with snow. The neutral transcript provides basic scientific knowledge on sunlight and the earth, and the opposition questions the correlation between gas emissions and geographic factors, thereby challenging simplistic attributions.
These examples illustrate how stance can be communicated through a combination of images and textual cues, which is crucial for training models to detect stance in multimodal data.

\begin{table}[ht!]
\centering
\renewcommand{\arraystretch}{1.3} 
\scriptsize 
\begin{tabular}{|>{\raggedright\arraybackslash}p{0.12\linewidth} 
                |>{\centering\arraybackslash}p{0.26\linewidth} 
                |>{\centering\arraybackslash}p{0.26\linewidth} 
                |>{\centering\arraybackslash}p{0.26\linewidth}|}
\hline
\textbf{Label} & \textbf{Example 1} & \textbf{Example 2} & \textbf{Example 3} \\
\hline

\textbf{Topic (Support)} & 
Climate Change: Mitigate or Adapt &
NASA's climate advisor discusses climate change &
Arctic Methane and Climate Change \\
\hline

\textbf{Image (Support)} & 
\begin{minipage}[c][2.0cm][c]{\linewidth}
  \centering
  \vspace*{2pt} 
  \includegraphics[width=0.7\linewidth]{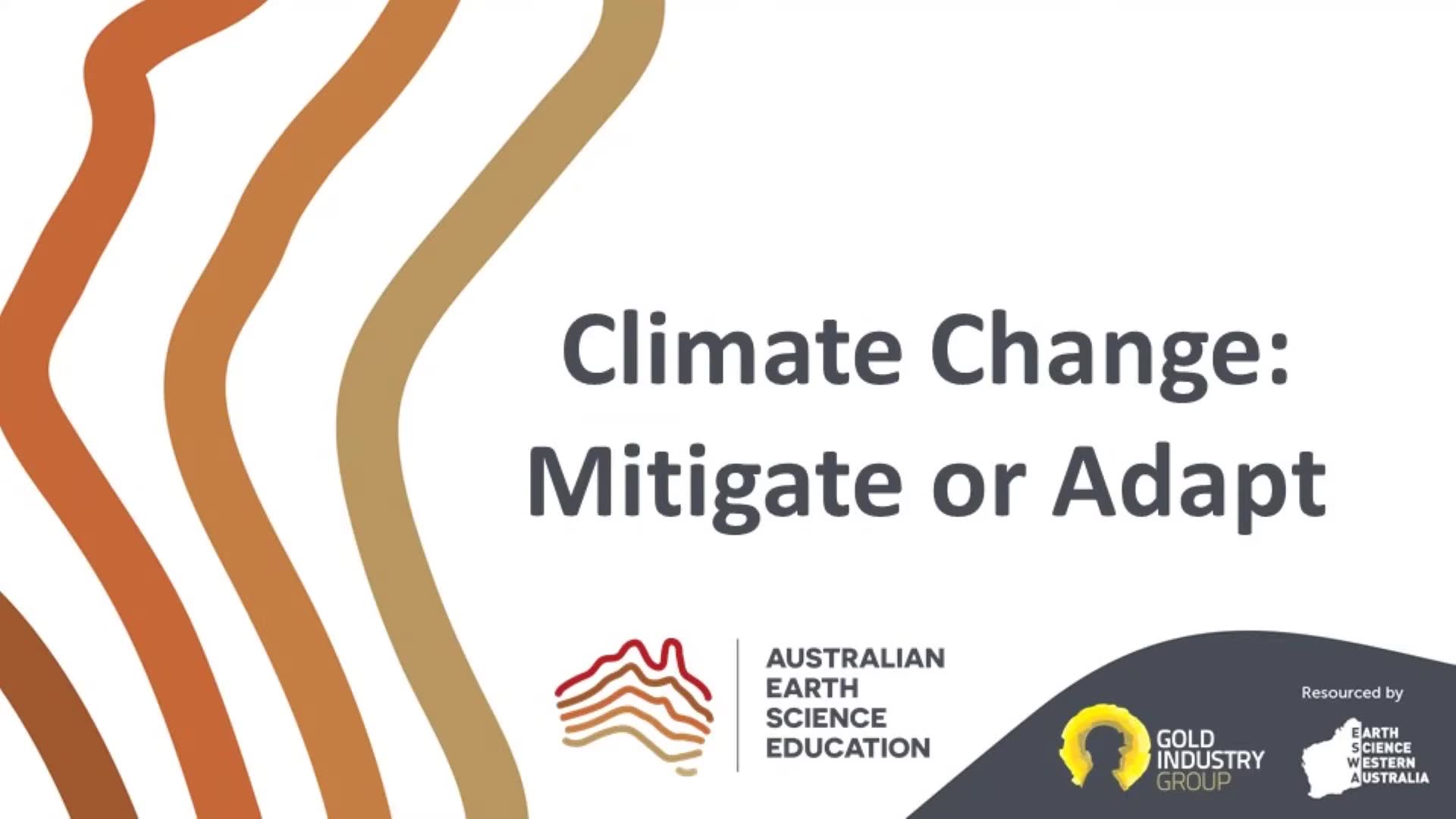}
  \vspace*{2pt} 
\end{minipage} &
\begin{minipage}[c][2.0cm][c]{\linewidth}
  \centering
  \vspace*{2pt} 
  \includegraphics[width=0.7\linewidth]{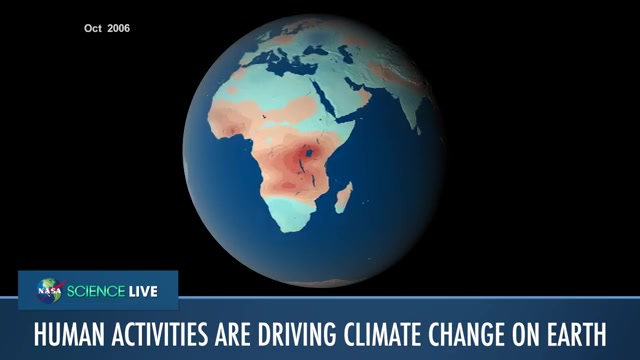}
  \vspace*{2pt} 
\end{minipage} &
\begin{minipage}[c][2.0cm][c]{\linewidth}
  \centering
  \vspace*{2pt} 
  \includegraphics[width=0.7\linewidth]{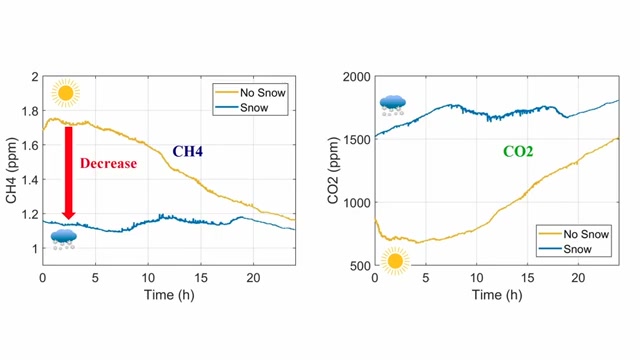}
  \vspace*{2pt} 
\end{minipage} \\
\hline

\textbf{Transcript (Support)} & 
\RaggedRight We are going to look at ways that people can mitigate or adapt to the changing climate and its flow-on effects. &
\RaggedRight Thanks to NASA Earth Science we know our planet and its climate are changing, and based on the evidence we also know human activities are driving this change. &
\RaggedRight Interestingly the methane concentration decreases when there is snow while the carbon dioxide concentration increases. \\
\hline

\textbf{Image (Neutral)} & 
\begin{minipage}[c][2.0cm][c]{\linewidth}
  \centering
  \vspace*{2pt} 
  \includegraphics[width=0.7\linewidth]{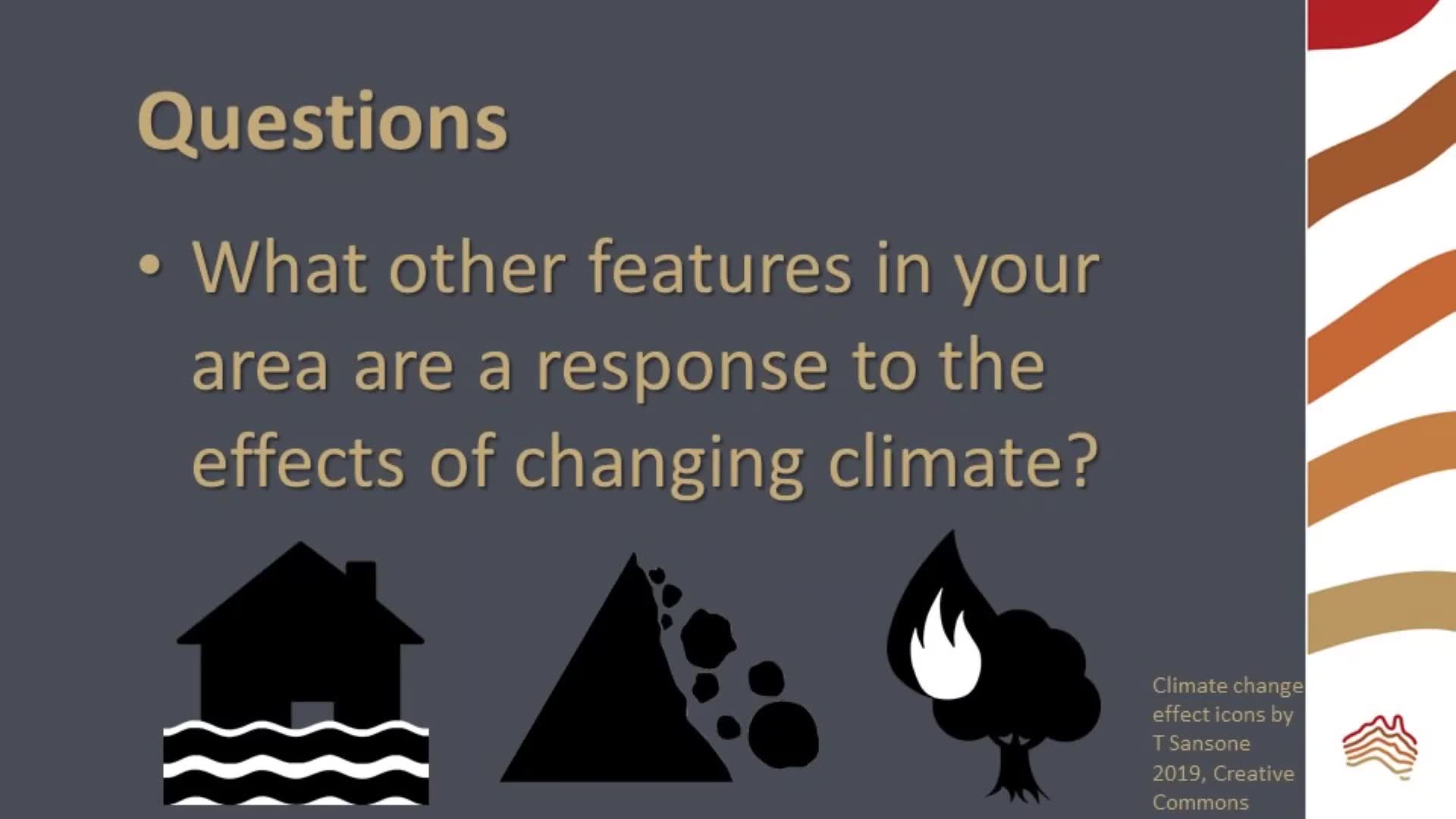}
  \vspace*{2pt} 
\end{minipage} &
\begin{minipage}[c][2.0cm][c]{\linewidth}
  \centering
  \vspace*{2pt} 
  \includegraphics[width=0.7\linewidth]{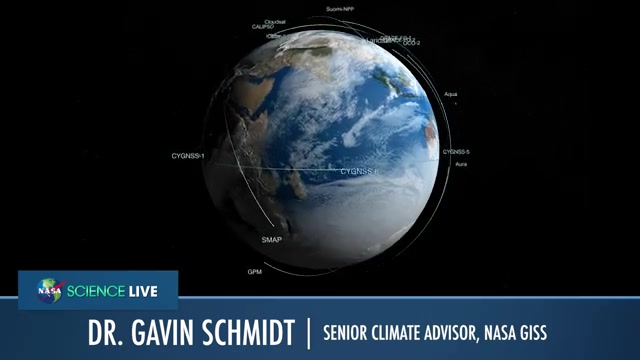}
  \vspace*{2pt} 
\end{minipage} &
\begin{minipage}[c][2.0cm][c]{\linewidth}
  \centering
  \vspace*{2pt} 
  \includegraphics[width=0.7\linewidth]{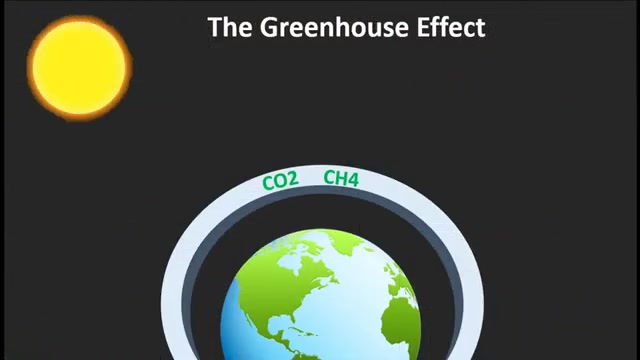}
  \vspace*{2pt} 
\end{minipage} \\
\hline

\textbf{Transcript (Neutral)} & 
\RaggedRight What other features in your local area are response to the effects of changing climate? &
\RaggedRight I've been a climate scientist with NASA for more than 25 years.  &
\RaggedRight Energy comes from the sun when sunlight hit the earth.   \\
\hline

\textbf{Image (Oppose)} & 
\begin{minipage}[c][2.0cm][c]{\linewidth}
  \centering
  \vspace*{2pt} 
  \includegraphics[width=0.7\linewidth]{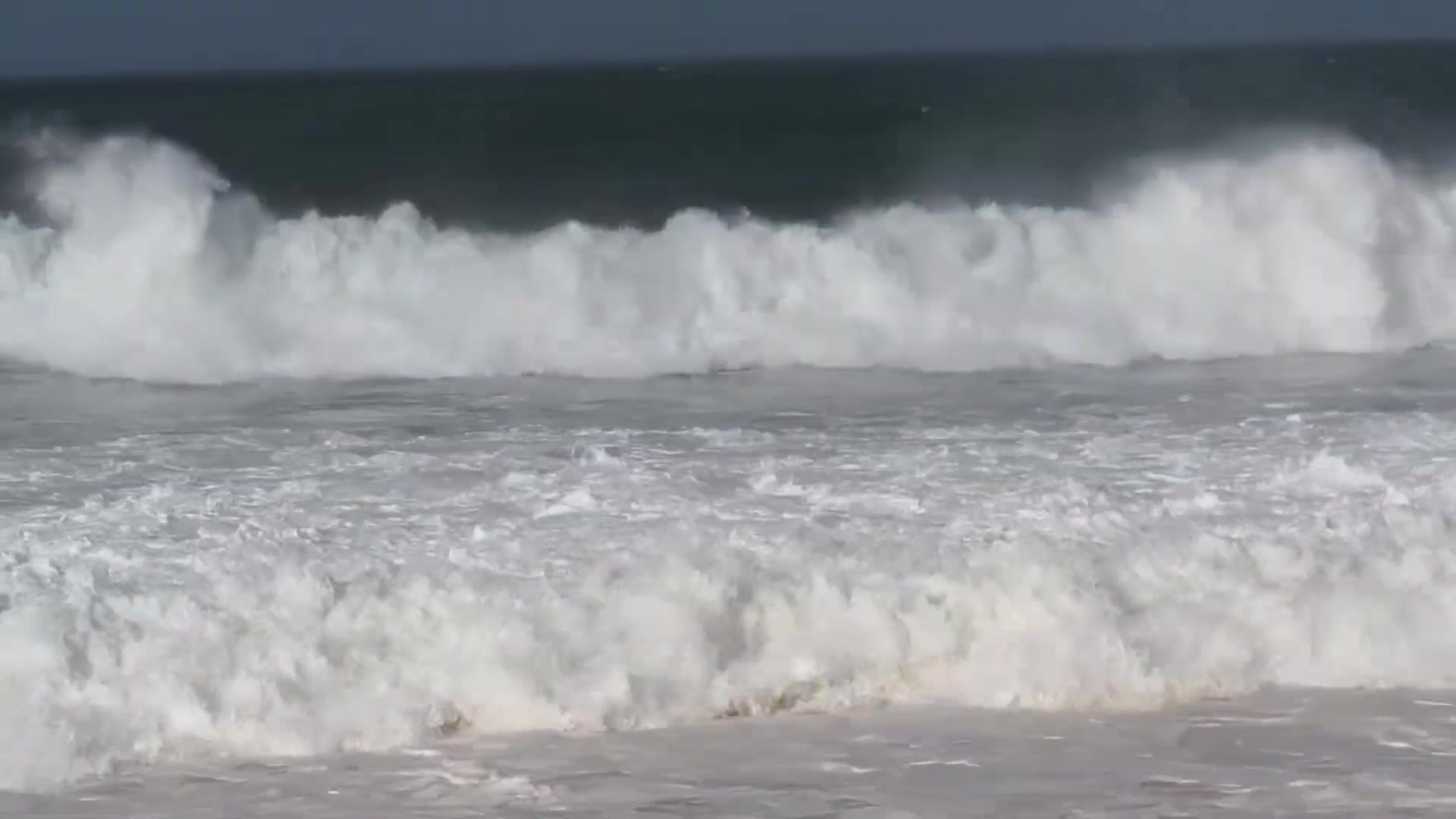}
  \vspace*{2pt} 
\end{minipage} &
\begin{minipage}[c][2.0cm][c]{\linewidth}
  \centering
  \vspace*{2pt} 
  \includegraphics[width=0.7\linewidth]{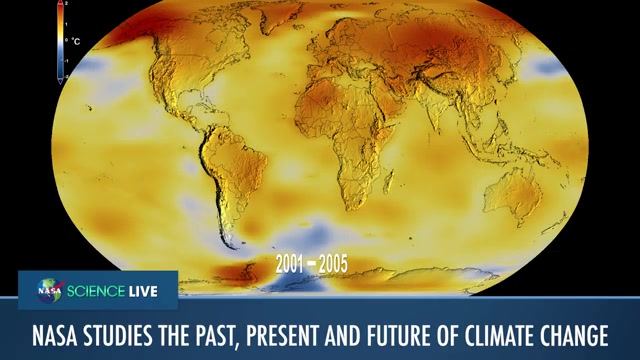}
  \vspace*{2pt} 
\end{minipage} &
\begin{minipage}[c][2.0cm][c]{\linewidth}
  \centering
  \vspace*{2pt} 
  \includegraphics[width=0.7\linewidth]{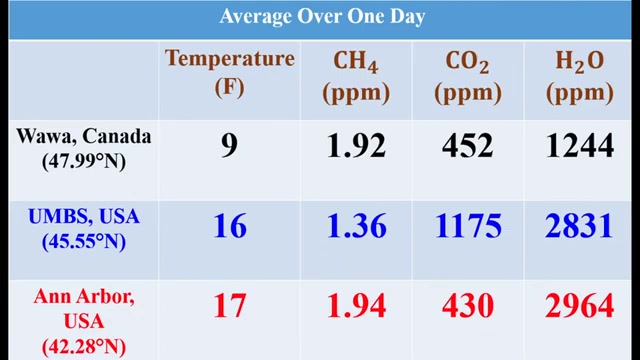}
  \vspace*{2pt} 
\end{minipage} \\
\hline

\textbf{Transcript (Oppose)} & 
\RaggedRight Even if we stop any further warming we have to deal with some of the effects we are already experiencing such as increased storms fires and heat. &
\RaggedRight It is a global problem felt on local scales and one that will be around for decades and centuries to come. &
\RaggedRight In summary there is no obvious dependence on latitude for methane and carbon dioxide emissions according to this one day average. \\
\hline
\end{tabular}
\caption{Examples Showing Alignment Between Video Frames And Corresponding Transcript Sentences Across Different Stance Categories}
\label{tab:stance_examples}
\end{table}

In today's digital landscape, where discussions about important societal issues \cite{Upadhyaya_Fisichella_Nejdl_2023,10.1145/3543873.3587643,MARTINEZ2023103294,rehman2025implihatevid} has increasingly occur through multimodal content combining text and images, traditional unimodal approaches to stance detection face significant limitations . While substantial progress has been made in text-based stance detection using advanced language models, these methods often fail to capture the rich, complementary information conveyed through visual elements such as infographics, protest imagery, or memes that frequently accompany social media posts.
The challenge of multimodal stance detection lies in effectively integrating heterogeneous data types without compromising computational efficiency. Even within existing multi-modal frameworks, the approaches commonly employ simple fusion of text and image features or use large pre-trained vision-language models that may not be optimized for the intricate task of stance interpretation. Moreover, existing methods frequently overlook the contextual relationships between different components of social media posts such as the interplay between an original post, its replies, and accompanying images, which can provide crucial signals for accurate stance classification.

Our paper presents a novel multimodal stance detection framework that addresses these challenges through several key innovations. Our approach combines the strengths of large language models for text understanding with efficient visual feature extraction and a novel joint text modeling technique that captures cross-modal interactions. 
The key contributions of our paper are as follows:

\begin{itemize}
    \item  We propose a multimodel approach combining textual and visual information for stance detection through hierarchical feature extraction and alignment. Our approach goes beyond simple feature concatenation by modeling deep interactions between modalities.

    \item  We introduce a two-stage context extraction process using (a) LLM-based summarization for distilling key arguments from text, and (b) domain-aware image caption generation for extracting stance-relevant visual information.

    \item  We develop a specialized transformer-based module that jointly processes text and generated image captions to captures complex cross-modal relationships.

\end{itemize}

The rest of the article is structured as follows. Section \ref{sec:related_work} presents an overview of related work in the domain. 
Section \ref{methodology} details the proposed methodology. Section \ref{sec:exp_ev} describes the datasets employed in our experiments, along with the experimental setup and baseline methods, and reports the results and provides an in-depth analysis. Section \ref{sec:discussion} discusses the theoretical and practical implications of our findings. Finally, Section \ref{sec:conclusion} summarizes the paper and discusses possible future direction.

\section{Related Work}\label{sec:related_work}
\subsection{Text-Based Stance Detection}
Zhang \textit{et al.} \cite{10.1145/3533430} utilises a Bidirectional Encoder Representations from Transformers (BERT) to encode the semantic representations of posts and their replies. The contextual embeddings are then integrated into a stance-centred graph structure. Dadas \textit{et al.} \cite{dadas-etal-2020-evaluation}  demonstrate significant improvements by applying BERT's contextual embeddings to capture stance expressions in social media text, achieving state-of-the-art results on political discourse. Work by Chen \textit{et al.} \cite{10651178} enhance traditional BERT models by incorporating sentiment-filtered knowledge graphs to improve stance prediction accuracy, demonstrating how auxiliary textual signals can boost performance. Similarly, Cheng \textit{et al.} \cite{CHENG2025113399} advance transformer architectures through contrastive heterogeneous topic graphs that model complex stance relationships in discussion threads. For low-resource scenarios, Yan \textit{et al.} \cite{10654680} propose an LLM-powered solution that generates synthetic training data via prompt engineering and knowledge distillation. While these methods showcase sophisticated text processing but they all operate exclusively on textual inputs. This represents a critical limitation, as real-world stance expression increasingly relies on multimodal content where visual elements such as memes, infographics, or protest imagery often carry decisive contextual cues~\cite{DANGI2025110895}. Our work directly addresses this gap by introducing the JTMo fusion module that jointly processes and aligns textual and visual stance signals, enabling a more holistic understanding.

\subsection{Multimodal Approaches for Stance Analysis}
Recent research has made valuable progress in multimodal stance detection through diverse approaches. Kuo \textit{et al.} \cite{10.1145/3589335.3651467} employ text-image concatenation with BERT and ResNet features, demonstrating the potential of combining modalities. Sengan \textit{et al.} \cite{10120685} innovatively incorporate tweet text, images, and stock signals via LSTM fusion, showing how auxiliary data can enrich stance analysis. Khiabani \textit{et al.} \cite{10098547} effectively utilise CLIP embeddings for few-shot adaptation, highlighting the versatility of pretrained multimodal representations. While these approaches establish important foundations for multimodal analysis, opportunities remain for enhanced cross-modal interaction modelling and domain specialisation. Building upon these contributions, our work introduces a hierarchical attention mechanism that dynamically aligns relevant image regions with textual arguments, enabling more precise interpretation of image-text relationships. We further integrate domain-aware prompting to guide feature extraction towards climate-specific cues, enhancing contextual relevance. Additionally, the proposed architecture is designed to be parameter-efficient, delivering strong performance without relying on external signal integration. This enables our model to better interpret complex multimodal stance expressions where visual evidence and textual claims interact through refined, context-sensitive pathways.

\subsection{Stance Detection using LLMs}
Recent work has explored various applications of large language models for stance detection. Lan \textit{et al.} \cite{Lan_Gao_Jin_Li_2024} demonstrate how multiple LLM agents with specialised roles can improve stance analysis through simulated discussions, though it relies entirely on textual inputs without considering visual context. Zhang \textit{et al.} \cite{zhang-etal-2024-llm-driven} effectively utilise external knowledge bases to enhance LLM performance on unseen targets, yet it processes knowledge in a unimodal textual format. Yang \textit{et al.} \cite{10.1145/3716856} combine social context with rumour detection through an innovative MIL framework, but does not incorporate the visual evidence often accompanying rumours.  Gambini \textit{et al.} \cite{10.1007/s10994-024-06587-y} provide comprehensive benchmarks of LLMs on Twitter data, identifying their strengths and limitations in textual stance classification. Chen \textit{et al.} \cite{app15115809} construct heterogeneous networks to model stance relationships, while Shafiei \textit{et al.} \cite{11006205} augment LLMs with external knowledge to improve contextual understanding, though both approaches focus solely on text-based social networks. While these works make valuable contributions to LLM-based stance analysis, they share common limitations in handling multimodal inputs and domain-specific adaptation. Our work advances beyond these approaches by integrating visual modality processing through a novel joint text-visual modelling framework, while maintaining the strengths of LLM-based contextual understanding. Where existing methods excel at textual analysis, our model additionally captures crucial visual stance cues prevalent in modern social media discourse, particularly for complex domains such as climate change, where visual evidence frequently complements textual arguments. The proposed architecture achieves this through domain-aware multimodal fusion while preserving the interpretability and generalisation capabilities demonstrated by these LLM-based approaches.

\section{Methodology}\label{methodology}
\begin{figure*}[!h]
    \centering
    \captionsetup{justification=justified}
    \includegraphics[width=16cm, height=7cm]{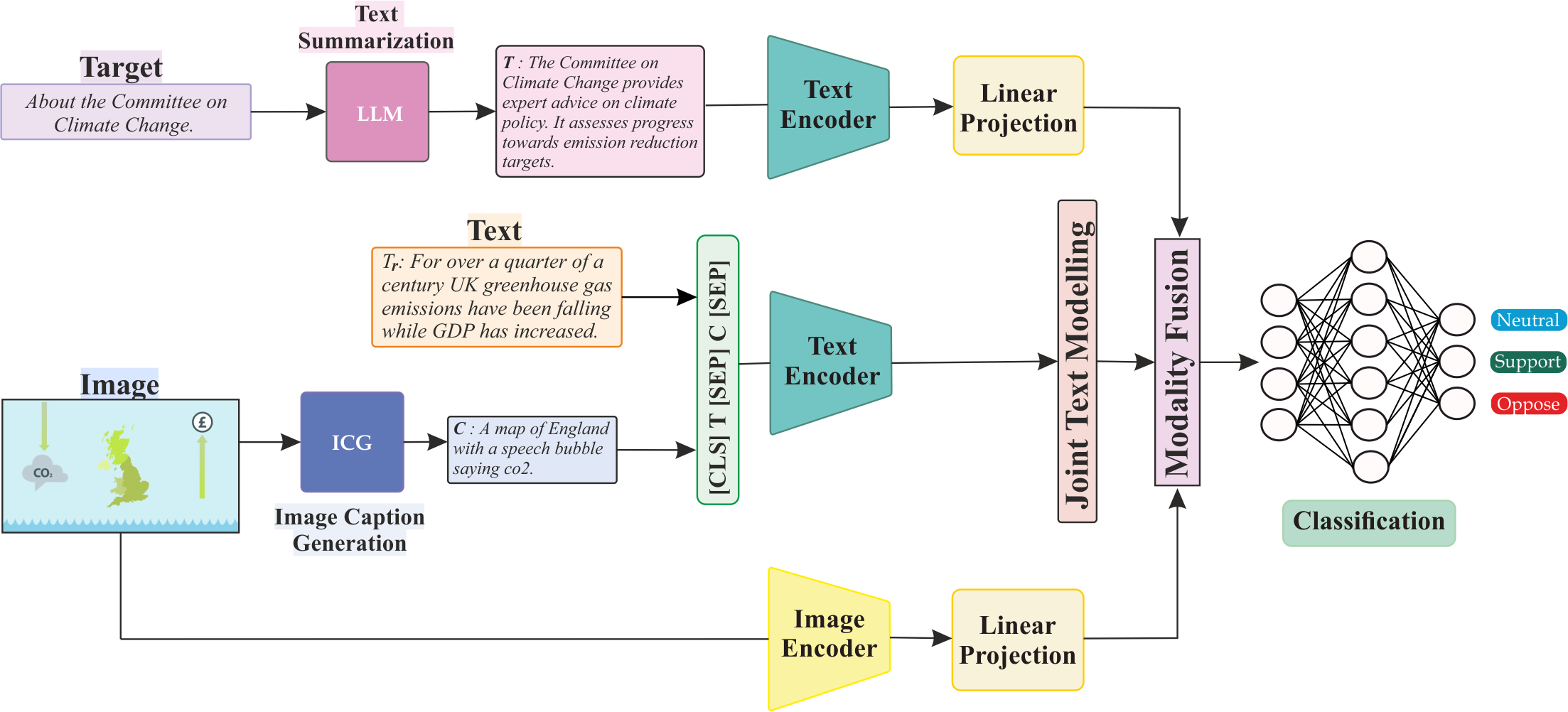}
  
    \caption{Proposed MultiModal Stance Detection Method}
    \label{fig:architecture}
\end{figure*}

This section describes the proposed multimodal framework for stance detection that jointly leverages textual and visual modalities. The architecture illustrated in Figure~\ref{fig:architecture} comprises four main components: (1) Stance-Relevant Context Extraction, (2) Feature Extraction,
(3) Joint Text Modelling, (4) Multimodal Feature Encoding, and (5) Classification.

Initially, in the Stance-Relevant Context Extraction, the source text is passed through a LLM to generate a context-rich summary. Simultaneously, the associated image is processed by an ICG model to extract a relevant caption or semantic description, aiding in the interpretation of visual implications related to climate change.
The summarized source text, reply text, and generated description are tokenized and encoded using a transformer-based text encoder. The image caption and reply text are further processed through a joint text modelling technique.
Meanwhile, the original image is encoded using a deep vision model. 
All resulting representations are linearly projected into a shared embedding space.
The summarized source text and image are directly fed into the model to form a unified multimodal representation.
This joint representation is then passed through a classification layer to predict the stance as one of the following categories: Support, Oppose, or Neutral.

\subsection{Stance-Relevant Context Extraction}
The first stage of our framework focuses on extracting meaningful context that is directly relevant to stance detection from both textual and visual modalities. This process ensures that only the most pertinent information is passed forward to the stance classification model, thereby improving interpretability and performance. This is achieved through two parallel modules:
\begin{enumerate}
    \item \textbf{Textual Context Extraction Using a Large Language Model (LLM):}
    The textual component of the input, typically a social media post, user comment, or news article, is processed using a pretrained Large Language Model (LLM) Mistral 7B model. The objective is to generate a concise summary that captures the core arguments or sentiments expressed in the source text.

For example, a tweet discussing climate change may be verbose or ambiguous, especially when the topic is specific, such as ``Big Data and Climate Change”. The LLM processes the text and condenses it into a focused summary such as:

``Big data is being used to model and understand climate change. It aids in predicting future trends and impacts of global warming”.

This summarization plays a crucial role in filtering out irrelevant content and highlighting the core message. By isolating the main viewpoint, it enables the stance detection module to more effectively interpret the author's position with respect to the topic.

     \item \textbf{Visual Context Extraction Using Image Caption Generation (ICG):}
     In parallel, if the post contains an image, we employ an Image Caption Generation (ICG) Prompt cap to interpret the visual content. The ICG module is guided by a domain-specific query tailored to the topic. For instance, in the context of climate change, the query might be:

“Generate the caption for the image”.

The ICG model then generates a caption or semantic interpretation 
$I_c$ that encapsulates the image's message, such as:

``A graphic with a sign saying the committee on climate change advances progress towards emission reduction target”

``A woman sitting at a table with a sign saying women in fisheries are vital to climate adaptation strategies”

``A globe with a smiley face on it with climate change is connected to many other global issues”

By asking targeted questions, the model ensures that the extracted visual information aligns with the domain context and supports the textual interpretation.
\end{enumerate}
\subsection{Feature Extraction}

For effective feature extraction from multimodal inputs, we employ a combination of transformer-based architectures tailored to textual and visual data. Specifically, BERT (Bidirectional Encoder Representations from Transformers) \cite{devlin-etal-2019-bert} is utilized to extract rich contextual embeddings from the social discussion text components, including the summarized source text, the reply text, and the image captions. BERT’s pretrained language model effectively captures the semantic and syntactic tones of these textual inputs, generating embeddings that serve as robust representations for downstream stance detection.

In parallel, the Visual Transformer (ViT) architecture is employed to extract discriminative features directly from the raw images associated with the posts. ViT processes the visual content by dividing images into patches and encoding these as sequences analogous to tokens in text, thereby enabling the application of transformer attention mechanisms to visual data. The features extracted by ViT encapsulate important visual patterns and contextual cues, which complement the textual embeddings.

By combining BERT-based textual features with ViT-based visual features, the model achieves a comprehensive multimodal representation that leverages both language and image modalities for improved stance understanding and classification.

\subsection{Joint Text Modelling}

In multimodal social media discourse, both textual and visual elements play a critical role in conveying the stance of a user toward a specific topic \cite{DAR2025125337, Bansal2025, multimodal,rehman2025multimodal,dar2025explainable}. Typically, the source text, such as a post or article, is summarized using a large language model to serve as a concise and focused representation of the target or topic under discussion. Simultaneously, the reply text is often accompanied by an image that serves as a visual cue, providing additional context or emotional undertones \cite{10.1145/3618057}. However, directly processing raw images alongside text poses challenges due to the heterogeneous nature of these modalities.

When the image caption aligns with the semantics of the reply, it may reinforce the stance expressed in the text. Conversely, if the caption contrasts with the text or the summarized target, it may indicate an opposing or skeptical stance. This interplay between image and text forms the foundation for the Joint Text Modelling (JTMo) framework, where both modalities are encoded together to enhance the model’s ability to detect tones, stance-relevant signals in multimodal conversations.
The JTMo input comprises two main components: (i) the social discussion source text, denoted as \(T\), and (ii) the image caption, denoted as \(C\). The caption serves as a concise semantic summary of the associated image, capturing relevant visual cues that complement or clarify the discussion. This approach bridges the modality gap by transforming images into domain- and context-aware textual descriptions aligned with the thematic concerns of the conversation \cite{10887864}.
The rationale for using image captions instead of raw images or standalone opinions is twofold. First, images can often contain noise or irrelevant visual details that may hinder effective stance detection if processed directly. By generating concise image captions, we distill the essential semantic information relevant to the discussion, reducing noise and improving focus. Second, concatenating the image caption with the target text and applying the transformer's multi-head attention mechanism allows the model to dynamically attend to the most pertinent textual and visual cues. This joint representation facilitates a coherent integration of modalities, enabling the model to capture subtle, contextually important interactions that enhance stance detection performance.

The input sequence is tokenized and concatenated following a BERT-style format:

\begin{equation}
\text{Input} = [\text{CLS}], t_{r1}, t_{r2}, \dots, t_{rn}, [\text{SEP}], c_1, c_2, \dots, c_m, [\text{SEP}]
\label{eq:concat}
\end{equation}

Here, \(\{t_{ri}\}\) represents tokens from the reply text \(T\), and \(\{c_i\}\) represents tokens from the image caption \(C\). Special tokens \texttt{[CLS]} and \texttt{[SEP]} distinguish the segments, preserving their contextual boundaries.
The embedded input is then processed through a stack of \(J\) transformer layers, each comprising multi-head self-attention and position-wise feedforward networks. For layer \(j\), the computations are:

\begin{align}
H_i^{(j)} &= \text{Attention}(Q_i, K_i, V_i) \\
\text{MultiHead}(H^{(j)}) &= \text{Concat}(H_1^{(j)}, \dots, H_h^{(j)}) W^O \\
\tilde{H}^{(j)} &= \text{LayerNorm}(H^{(j-1)} + \text{MultiHead}(H^{(j-1)})) \\
H^{(j)} &= \text{LayerNorm}(\tilde{H}^{(j)} + \text{FFN}(\tilde{H}^{(j)}))
\end{align}

where, \(Q_i\), \(K_i\), and \(V_i\) are the query, key, and value matrices of the \(i\)-th attention head, and \(W^O\) is the learned output projection matrix. The self-attention mechanism models pairwise token interactions across both text and caption, enabling the model to dynamically focus on salient multimodal features. Residual connections and layer normalization promote stable training and effective gradient flow, while the feedforward network enhances representation capacity.

After \(J\) layers, the model outputs the final contextual embedding \(H^{(J)}\), referred to as the \textit{Text-Social-Visual} embedding \(H'\), which jointly encodes semantic information from both the social text and the aligned image caption given in Equation \ref{eq:stack}.

\begin{equation}
H' = H^{(J)}
\label{eq:stack}
\end{equation}

By jointly modeling textual and visual semantics, JTMo captures subtle multimodal interactions, such as visual evidence that reinforces or contradicts textual claims, interactions often missed by text-only models.
Additionally, exploiting a BERT-style transformer backbone enables JTMo to benefit from large-scale pretraining, enhancing its robustness and generalizability across diverse topics and linguistic styles. The multi-head attention mechanism flexibly attends to the most relevant tokens across modalities, adapting to the varying importance of textual and visual cues for accurate stance detection \cite{DAR2024112526}.


\subsection{Multimodal Feature Fusion}

The output from the JTMo Encoder is concatenated with the individual textual and visual embeddings. The output representations from the {Joint Text-Modal (JTMo) Encoder} are fused with individual textual and visual embeddings to enable effective multimodal learning. Since the textual ($H_t$) and visual ($H_v$) features originate from different modalities and encoders (BERT and ViT, respectively), it is essential to project them into a {common embedding space} before fusion.
To ensure that features from different modalities are compatible and comparable, we use linear projection layers to map the representations into a shared $d$-dimensional embedding space:

\begin{equation}
\tilde{H}_t = W_t H_t + b_t
\label{eq:text_proj}
\end{equation}

\begin{equation}
\tilde{H}_v = W_v H_v + b_v
\label{eq:visual_proj}
\end{equation}

\begin{equation}
\tilde{H}_j = W_j H_j + b_j
\label{eq:jtmo_proj}
\end{equation}

where, $H_t \in \mathbb{R}^{n \times D_t}$ is the Textual representation from BERT, $H_v \in \mathbb{R}^{N \times D_v}$ is the Visual representation from ViT, $H_j \in \mathbb{R}^{1 \times D_j}$ is the Output of JTMo Encoder (global interaction features), $W_t, W_v, W_j$ are Learnable projection matrices mapping inputs to $\mathbb{R}^{d}$, $b_t, b_v, b_j$ are Bias terms, $\tilde{H}_t, \tilde{H}_v, \tilde{H}_j \in \mathbb{R}^{d}$ are Projected representations.

After projection, the representations are concatenated to form a unified multimodal representation:

\begin{equation}
H_{\text{fused}} = [\tilde{H}_t \, \| \, \tilde{H}_v \, \| \, \tilde{H}_j]
\end{equation}

where, $\|$ denotes concatenation.
This fused representation $H_{\text{fused}} \in \mathbb{R}^{3d}$ combines Linguistic context (from BERT), Visual semantics (from ViT), and Cross-modal interaction (from JTMo Encoder).
Different modalities (text and image) typically have different statistical properties and dimensionalities such as BERT outputs a 768-dimensional embedding per token, ViT outputs a global 768-dimensional embedding from the [CLS] token or patch embeddings, and JTMo may have its own dimensionality depending on fusion depth.
Directly concatenating them without projection would lead to misalignment, unstable training, and sub-optimal performance. By mapping all features into the same semantic space, we ensure all features contribute equally. Allowing similarity computations, attention mechanisms, and reducing parameter overhead improves fusion.

\subsection{Classification}
The model creates a joint encoded representation of the textual and visual modalities. These representations are then aligned and combined into a single feature vector and finally classified into three-class probabilities. The image caption representation, pooled contextual features, and summarized source text embedding are all linearly projected into a common 512-dimensional space.  Effective fusion is made possible by this dimensional alignment, which maintains the semantic granularity of each modality while allowing for consistent comparison between them.  The $H_{fused}$ is a comprehensive multimodal feature vector created by concatenating the projected representations.  This combination provides a rich foundation for classification later on by capturing subtle relationships between the text, visual cues, and contextual signals.
A dropout layer is applied at this stage to mitigate overfitting by introducing controlled regularization during training. By integrating these heterogeneous signals, the model is better equipped to interpret stance-bearing cues that emerge from cross-modal interactions.

The final fused vector 
$H_{fused}$ is passed through a linear transformation layer that functions as the classification head, mapping the integrated multimodal representation to a predefined set of three stance categories: Support, Oppose, and Neutral. This classification head is trained to identify the underlying stance by exploiting the semantic features distilled from the textual content, contextual information, and visual cues. A dropout layer is applied prior to this classification step to improve generalization and reduce overfitting, thereby encouraging the model to learn robust and discriminative features across modalities. The use of dedicated linear projections for each modality before fusion ensures that all inputs suchas 
 text, pooled contextual signals, and image captions contribute uniquely and effectively to the fused representation. This fusion mechanism allows the model to mitigate uncertainties or ambiguities that may arise when a single modality lacks sufficient stance-related cues. By incorporating complementary information from both the textual and visual domains, the model achieves a more holistic understanding of the user's stance. The integrated classification pipeline thus plays a vital role in enabling accurate and context-aware stance inference, especially in complex or visually grounded discussions on social media platforms.

\section{Experimental Evaluations}\label{sec:exp_ev}
The datasets utilized for the experimental assessment of our proposed framework for stance categorization is presented in this section. We describe the experimental design in detail below, along with the comparison techniques and compare the performance with state-of-the-art methods.
\subsection{Dataset}
The {MultiClimate} dataset \cite{wang-etal-2024-multiclimate} is a multimodal dataset specially designed for stance detection on climate change content, incorporating both textual and visual information. It consists of 100 Creative Commons licensed YouTube videos related to climate change, which are segmented into 4,209 frame transcript pairs. Each data instance combines a single video frame with a corresponding transcript sentence, labeled with one out of the three stance labels: \textit{Support}, \textit{Neutral}, or \textit{Oppose}. The dataset is split into training, development, and test splits, maintaining a balanced distribution of stances across all splits.
The stance labels are divided across the train, development, and test splits is presented in Table~\ref{tab:multiclimate_summary}.
By integrating linguistic cues from text and contextual cues from video frames, MultiClimate enables robust multimodal analysis for stance classification tasks.

\begin{table}[ht]
\centering
\begin{tabular}{lccccc}
\toprule
\textbf{Split} & \textbf{\# Videos} & \textbf{Support} & \textbf{Neutral} & \textbf{Oppose}  & \textbf{Total}\\
\midrule
Train & 80  & 1,449 & 1,036 & 887&3,372  \\
Dev   & 10  & 204   & 83    & 130 & 417 \\
Test  & 10  & 194   & 73    & 153  & 420\\
\midrule
\textbf{Total} & \textbf{100} & \textbf{1,847} & \textbf{1,192} & \textbf{1,170} & \textbf{4,209} \\
\bottomrule
\end{tabular}
\caption{Overview of the MultiClimate Dataset}
\label{tab:multiclimate_summary}
\end{table}

\subsection{Evaluation Metrics}

To assess the performance of the proposed stance detection model, we use standard evaluation metrics including {Accuracy}, {Precision}, {Recall}, and {F1-Score}.  

\textit{Accuracy:} Accuracy measures the proportion of correctly predicted instances out of the total number of instances and is defined in Equation~\ref{eq:accuracy}.

\begin{equation}
\text{Accuracy} = \frac{\sum_{i=1}^{N} TP_i}{\sum_{i=1}^{N} (TP_i + FP_i + FN_i)}
\label{eq:accuracy}
\end{equation}

\textit{Precision:} Weighted precision calculates the precision for each class and computes their average, weighted by the true instances in ea of the classes. It is given in Equation~\ref{eq:precision}.

\begin{equation}
\text{Precision} = \sum_{i=1}^{N} w_i \cdot \frac{TP_i}{TP_i + FP_i}
\label{eq:precision}
\end{equation}

\textit{Recall:} Weighted recall is the average recall across all classes, weighted by the number of true instances per class. It is given in Equation~\ref{eq:recall}.

\begin{equation}
\text{Recall} = \sum_{i=1}^{N} w_i \cdot  \frac{TP_i}{TP_i + FN_i}
\label{eq:recall}
\end{equation}

\textit{F1-Score:} The weighted F1-score is the harmonic mean of weighted precision and weighted recall, taking class imbalance into account. It is defined in Equation~\ref{eq:f1}.

\begin{equation}
\text{F1}= \sum_{i=1}^{N} w_i \cdot  \frac{2 \cdot \text{Precision}_i \cdot \text{Recall}_i}{\text{Precision}_i + \text{Recall}_i}
\label{eq:f1}
\end{equation}

Here, \( TP_i \), \( FN_i \), and \( FP_i \) represent the true positives, false negatives, and false positives for class \( i \); \( N \) is total classes, and \( w_i = \frac{n_i}{\sum_{j=1}^{N} n_j} \) is the weight of class \( i \), proportional to the number of true instances \( n_i \) in that class. These weighted metrics ensure a fair evaluation of model performance, especially in the presence of class imbalance across stance categories.

\subsection{Experimental Results}

This section presents the empirical evaluation of the proposed model on the MultiClimate \cite{wang-etal-2024-multiclimate} dataset. We compare its performance with following state-of-the-art models to demonstrate the perfromance and effectiveness.

\textit{BERT}  (Bidirectional Encoder Representations from Transformers) \cite{devlin-etal-2019-bert} is a pre-trained model for various NLP tasks. 
It uses a transformer architecture to learn contextual relationships between words in a text. 
BERT has set new performance benchmarks on a wide range of NLP tasks.

\textit{ResNet-50} (Residual
Network) \cite{resnet}  is a deep residual network with 50 layers designed for image classification. 
It introduces residual connections, which help mitigate the vanishing gradient problem. 
ResNet-50 has been successful in achieving high accuracy on large-scale datasets such as ImageNet.

\textit{ViT} (Vision Transformer) \cite{vit} is a model that applies transformer architectures to image data. 
It partitions images into fixed-size patches and processes them as sequences, similar to how transformers handle text. 
ViT has shown competitive performance, often surpassing traditional CNNs on image recognition tasks.

\textit{BLIP} (Bootstrapping Language-Image Pre-training) \cite{blip} is a vision-language pre-training model. 
It is designed to align images with textual descriptions to improve multimodal understanding. 
BLIP excels in tasks such as image captioning and visual question answering.

\textit{CLIP}  (Contrastive Language-Image Pre-training) \cite{clip} is a model that learns to relate images to textual descriptions. 
CLIP performs exceptionally well in zero-shot learning scenarios across various tasks.
\textit{Fakefind} \cite{fakefind} introduces a new system that identifies signs of depression during the COVID-19 period by analyzing diverse social media content. It uses not just text, but also images, user behavior, and linked external content to better understand emotional states. A specialized neural network processes image data to aid in prediction. The model shows improved results over prior methods and offers fresh insights through the use of a custom COVID-19 dataset.

\textit{MemeCLIP} \cite{shah-etal-2024-memeclip} 
 explores the effectiveness of CLIP's joint vision-language embeddings for the task of multimodal meme classification, which requires understanding both textual and visual content.
The model fine-tunes CLIP to align visual and textual cues, enabling improved detection of offensive, hateful, or misleading content in internet memes.
\begin{table}[ht]
\centering
\begin{tabular}{lccccc}
\toprule
\textbf{Model} & \textbf{Accuracy} &\textbf{Precision}&\textbf{Recall}& \textbf{F1} & \textbf{\# Params} \\
\midrule
CLIP \cite{clip}& 0.461 &0.213 &0.461&0.291 & 151.3M \\
BLIP \cite{blip} & 0.462 &0.213&0.462& 0.292 & 470M \\
Fakefind \cite{fakefind}&0.489	&0.239	&0.489	&0.324&	228.3M\\
BERT \cite{devlin-etal-2019-bert} + ResNet50 
 \cite{resnet}& 0.726 & 0.726&0.726&0.723 & 111.7M \\
BERT \cite{devlin-etal-2019-bert}+ ViT \cite{vit}& 0.731 & 0.740& 0.731& 0.732 & 196.8M \\
MemeCLIP \cite{shah-etal-2024-memeclip}&0.712&0.704&0.723&0.684&430M\\
\bottomrule
\textbf{Proposed Methodology}& \textbf{0.762} & \textbf{0.763}&\textbf{0.762} & \textbf{0.762}  & \textbf{202M} \\
\bottomrule
\end{tabular}
\caption{Performance Comparison Results on the MultiClimate Dataset}
\label{tab:multimodal_results}
\end{table}

Table \ref{tab:multimodal_results} The table presents a comprehensive comparison of various multimodal models evaluated on the MultiClimate \cite{wang-etal-2024-multiclimate} dataset in terms of Accuracy, Precision, Recall, F1 Score, and the number of parameters. These models integrate textual and visual modalities to determine the stance of social media content, with performance metrics indicating their effectiveness.
The baseline models, CLIP \cite{clip} and BLIP \cite{blip}, demonstrate relatively poor performance, with accuracy values of 0.461 and 0.462, and F1 scores of 0.291 and 0.292, respectively. Despite their large model sizes (151.3M and 470M parameters), these models are pretrained for general vision-language alignment tasks and are not specifically optimized for stance detection. This likely contributes to their low precision and recall values, as they may struggle to capture the intricate relationships between text and images that are critical for accurately identifying stance.
Fakefind \cite{fakefind} shows a slight improvement over CLIP and BLIP, with an accuracy of 0.489 and an F1 score of 0.324. While it incorporates both textual and visual information, the modest gains suggest that the model may still lack task-specific architectural enhancements or sufficient domain adaptation required for more effective stance inference.
Significant performance improvements are observed in models that utilize pre-trained language models such as BERT \cite{devlin-etal-2019-bert} in combination with visual encoders. The BERT + ResNet50 \cite{resnet} model achieves an accuracy of 0.726 and an F1 score of 0.723, indicating a strong alignment between its textual and visual components. Similarly, BERT + ViT \cite{vit} further improves the performance, attaining an accuracy of 0.731 and an F1 score of 0.732. The Vision Transformer (ViT) likely provides richer image representations than ResNet50 , which contributes to the boost in performance.
MemeCLIP \cite{shah-etal-2024-memeclip}, designed for meme and social media content understanding, also performs competitively with an F1 score of 0.684. Although its architecture is tailored for multimodal misinformation tasks, it still lags behind more integrated text-image fusion models such as BERT + ViT, suggesting limitations in its ability to generalize across stance categories.
Our proposed method achieve superior performance all other models across all evaluation metrics, achieving an accuracy and F1 score of 0.762. Notably, it achieves this performance with only 202M parameters, indicating a well-optimized architecture. The superior results may be attributed to its task-specific design, which likely incorporates advanced cross-modal fusion strategies, emotion-aware modeling \cite{PANGTEY2025111109}, or attention mechanisms \cite{REHMAN2025103895, RAGHAW2025107363} tailored to classification tasks. Its balanced performance across precision and recall further underscores its robustness in handling class distributions and content variability present in real-world multimodal data.
In conclusion, the results highlight the importance of domain-specific architectural enhancements over generic multimodal pretraining. The proposed methodology demonstrates how targeted model design can lead to more accurate and reliable stance detection, even with a moderate parameter count compared to larger, less specialized models.

\begin{figure}[htbp]
    \centering    \includegraphics[width=0.8\textwidth]{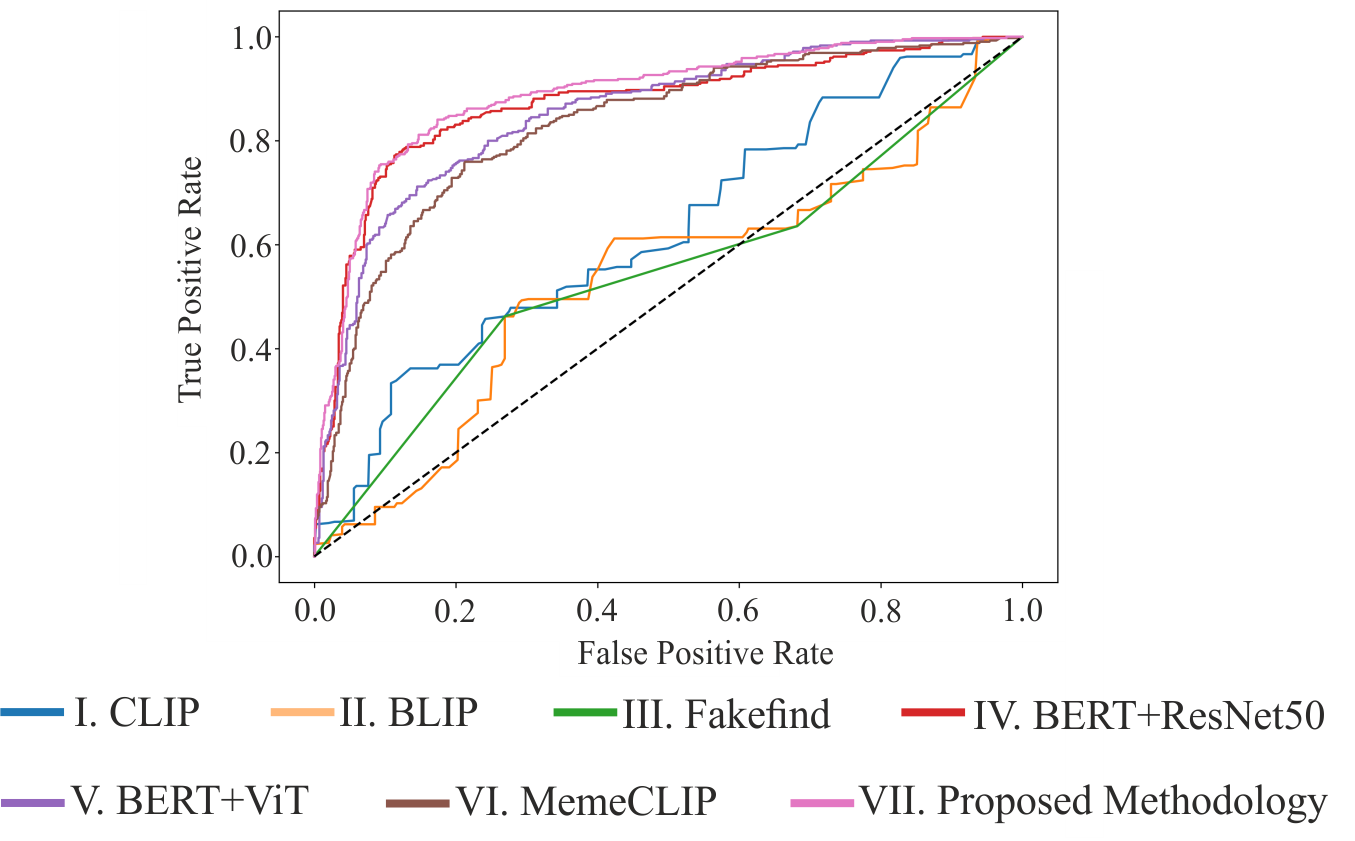}  \caption{ROC Curves With Weighted AUC For Baseline Models And The Proposed Method}

    \label{fig:roc}
\end{figure}
Figure \ref{fig:roc} demonstrates ROC curve which compares the performance of various multimodal models for the stance classification task using the Weighted AUC metric. The proposed method outperforms all others with a Weighted AUC of 0.88, indicating strong discriminative capability. In contrast, CLIP and BLIP show significantly lower performance, with AUCs of 0.53 and 0.44 respectively. Overall, models that combine textual and visual information, such as BERT+ResNet50 and BERT+ViT, perform considerably better than unimodal baselines. This superior performance highlights the effectiveness of multimodal integration, which forms the foundation of our proposed method.

\subsubsection{Ablation Analysis}
\begin{figure*}[!h]
    \centering
    \captionsetup{justification=justified}
    \includegraphics[width=11cm, height=8cm]{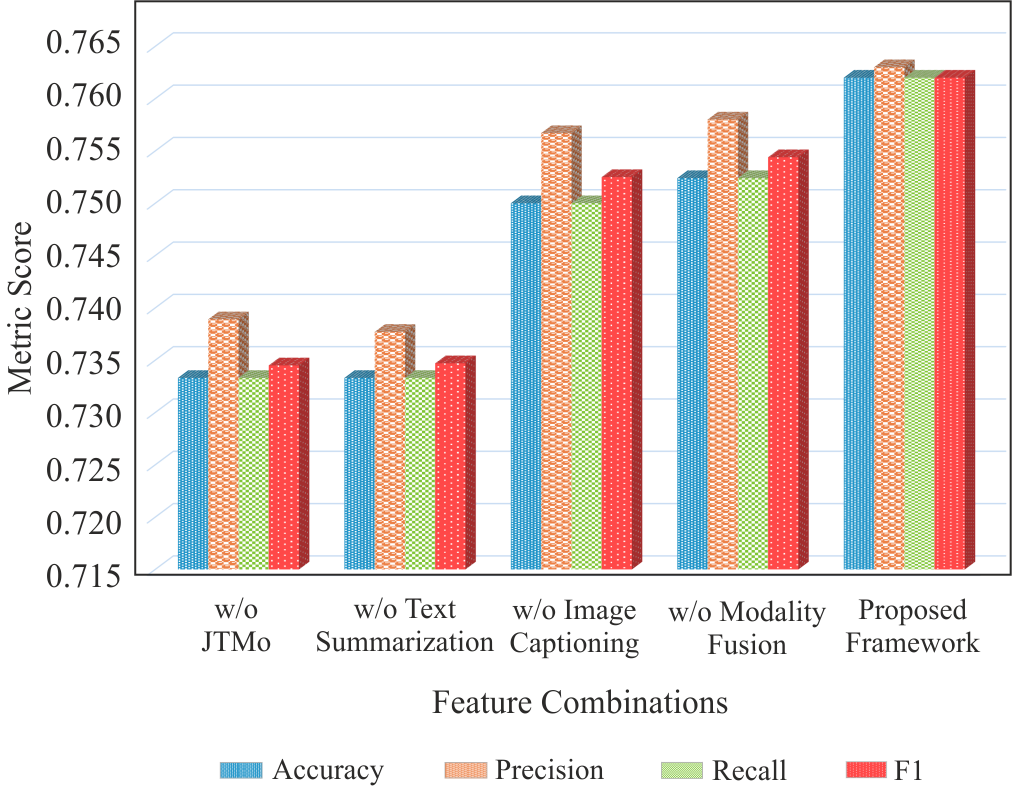}
  
    \caption{Ablation analysis results on MultiClimate dataset}
    \label{fig:ablation}
\end{figure*}

We outline
the detailed experimental analysis of our methodology with different modules such as w/o JTMo, w/o Text Summarization, w/o Image Captioning, w/o Modality Fusion and Proposed Methodology in this section.
The ablation analysis presented in Figure \ref{fig:ablation} evaluates the impact of various components of the proposed multimodal stance detection framework by systematically removing JTMo framework, text summarization, image captioning, modality fusion and observing changes in performance across four key metrics: Accuracy, Precision, Recall, and F1-score.
From the figure and the accompanying table, it is evident that the Proposed Framework, which integrates all components, achieves the best overall performance with an Accuracy, Precision, Recall, and F1-score of 0.762. This serves as the baseline for comparison. When the modality fusion mechanism is removed, performance slightly drops, with the F1-score decreasing to 0.754, highlighting the importance of effectively combining textual and visual information for accurate stance detection. A further decline is observed upon the removal of the image captioning module, which has an F1 score of 0.752, suggesting that visual semantics play a valuable role in contextual understanding.
The impact is even more pronounced when text summarization is excluded, reducing the F1-score to 0.7347. This implies that condensing lengthy textual content helps the model focus on essential information, enhancing its discriminative ability. Similarly, excluding the JTMo module results in a comparable drop to F1 score of 0.7345, underlining the significance of JTMo between modalities to achieve coherent multimodal representations.
Overall, the results clearly demonstrate that each component contributes incrementally to the model’s success. The consistent performance drop with each ablation reinforces the effectiveness of the full multimodal architecture, where textual and visual features are jointly optimized, semantically enriched, and coherently fused to enhance stance classification in social media content.

\subsubsection{Computational Complexity}
In deep learning, computational complexity refers to the amount of resources such as memory, processing power, and time required to train and run models~\cite{bansal2025retrieval,RAGHAW2025108225}. It is primarily influenced by the number of parameters in the model and the architectural depth. Models with higher parameter counts tend to achieve better performance but at the cost of increased inference time, greater memory consumption, and reduced scalability. Evaluating both model size (in terms of parameters) and inference time is essential for understanding the trade-offs between accuracy and efficiency.
\begin{figure}[htbp]
    \centering    \includegraphics[width=1\textwidth]{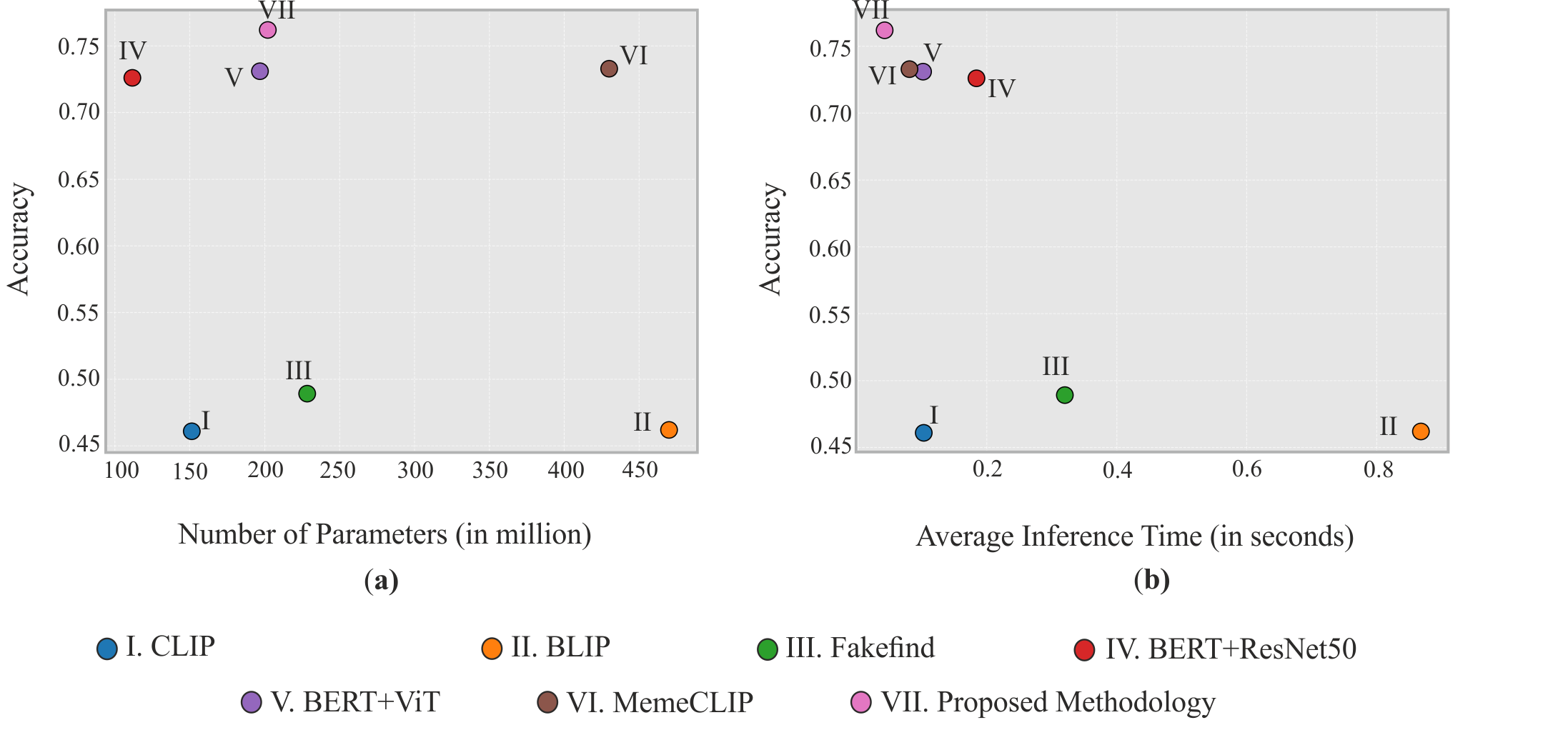} 
    \caption{Trade-Offs Among Parameter Count (In Millions), Average Inference Time (In Seconds), And Accuracy Across Methods. Circle Size Indicates Model Parameters, Highlighting Computational Cost And Efficiency.
}
    \label{fig:comp}
\end{figure}
Table \ref{fig:comp}(a) illustrates the relationship between model accuracy and the number of parameters (measured in millions), which serves as a proxy for inference efficiency or average inference time. Each data point represents a distinct model, allowing for a clear comparison of their performance-efficiency trade-offs. Models such as CLIP and BLIP possess relatively large parameter sizes (0.103M and 0.868M, respectively) but exhibit low accuracy values around 46\%, indicating that an increase in model complexity does not inherently guarantee improved predictive performance.
In contrast, multimodal models such as BERT + ResNet50, BERT + ViT, and MemeCLIP demonstrate higher accuracies (ranging from 72.6\% to 73.3\%), suggesting that integrating visual and textual modalities contributes significantly to improved model performance. However, these improvements come with a corresponding increase in computational overhead due to larger model sizes.
The most notable observation from the figure is the performance of the Proposed Methodology, which achieves the highest accuracy (76.2\%) while utilizing the fewest parameters (0.043M). This result underscores the efficiency and effectiveness of the proposed approach, which not only outperforms existing models in terms of predictive accuracy but also maintains a lightweight architecture suitable for deployment in resource-constrained environments. The balance achieved by the proposed model between accuracy and computational efficiency highlights its practical applicability in real-world settings where both performance and inference speed are critical.

Table \ref{fig:comp}(b) presents a comparative analysis of various multimodal models by plotting their classification accuracy against average inference time (in seconds). This visualization highlights the trade-off between predictive performance and computational efficiency across different approaches. Models such as CLIP and BLIP exhibit higher inference times (0.103s and 0.868s, respectively) while achieving relatively low accuracy of 46\%, indicating limited practical utility despite their larger computational footprints. Similarly, Fakefind \cite{fakefind}, though faster than BLIP, provides only marginal improvements in accuracy, suggesting that model optimization must go beyond parameter reduction.

On the other hand, BERT + ResNet50, BERT + ViT, and MemeCLIP achieve significantly better accuracy scores above 72\%, validating the importance of fusing textual and visual modalities. However, their inference times remain moderately high, which may limit their scalability in real-time or resource-constrained environments. The Proposed Methodology stands out by attaining the highest accuracy of 76.2\% while also achieving the lowest inference time of 0.043s, demonstrating a superior balance between efficiency and effectiveness. This result underlines the robustness and deployment readiness of the proposed approach, offering a promising solution for real-world applications where both rapid inference and high predictive accuracy are critical.

\section{{Discussion}}\label{sec:discussion}
\begin{figure}[htbp]
    \centering    \includegraphics[width=1\textwidth]{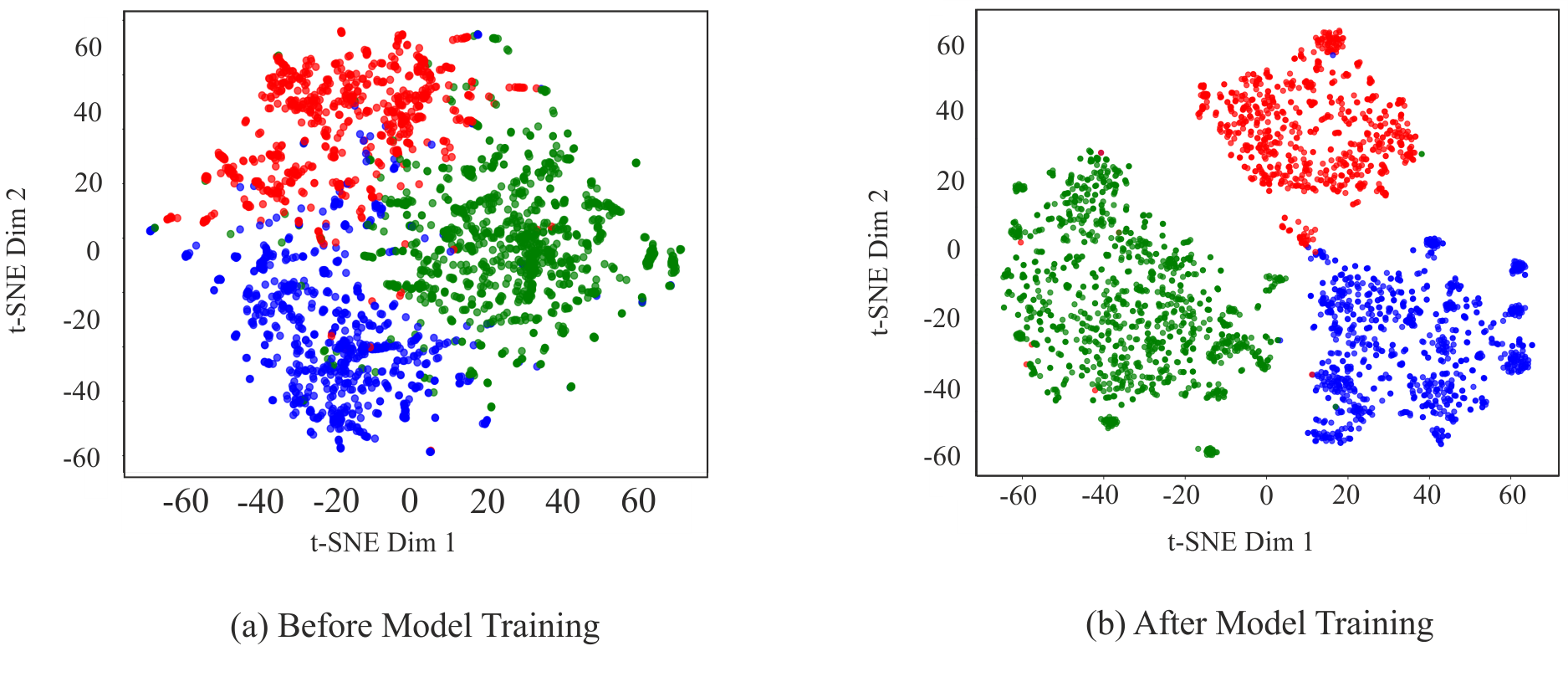}  \caption{t-SNE Visualization Before And After Model Training, Illustrating Improved Separability Among Stance Classes.
\textcolor{green}{Green} Indicates The \textit{Support} Class, \textcolor{red}{Red} Corresponds To The \textit{Oppose} Class, And \textcolor{blue}{Blue} Denotes The \textit{Neutral} Class.
}
    \label{fig:tsne}
\end{figure}

This research introduces a comprehensive framework for multimodal stance detection in social media content, yielding significant findings with implications for the broader field of natural language understanding. The proposed approach achieves a strong performance underscoring the efficacy of integrating textual and visual modalities in capturing user stance. Unlike unimodal methods or naive concatenation strategies, our framework demonstrates that a carefully constructed fusion strategy can enhance classification performance by exploiting the complementary nature of different data modalities.
Central to the framework's effectiveness is the joint modeling module, which enables an understanding of the interplay between textual content and associated imagery. This module facilitates the alignment of semantic signals across modalities, allowing the model to better interpret posts where the stance is implied through both language and visual cues. The use of large language model (LLM)-based summarization plays a pivotal role in condensing verbose and noisy user-generated text into its most stance-relevant elements. By extracting the essential argumentative components of the text, the model can focus on the core stance-bearing content without being distracted by extraneous information common in social media discourse.
Additionally, the inclusion of domain-specific image captioning contributes significantly to the model's performance. By generating captions tailored to the climate change domain, the system is better equipped to identify visual elements pertinent to the discussion at hand. This approach ensures that the visual modality is not treated generically, but instead contributes targeted information aligned with the stance detection objective.
The effectiveness of the proposed multimodal learning approach is further illustrated in Figure \ref{fig:tsne}, which presents a t-SNE visualization of the learned feature space before and after model training.
Nonetheless, several limitations of the current work must be acknowledged. The framework is presently limited to English-language content, which restricts its utility in multilingual or global social media environments. Furthermore, while the current design focuses on text-image pairs, social media platforms increasingly feature a wide range of content formats, including videos and audio, which may carry additional stance indicators. The absence of mechanisms to address sarcasm, irony, and other complex linguistic phenomena further presents challenges for robust stance interpretation in informal online discourse.
\section{Conclusion}\label{sec:conclusion}
The paper presents an effective multimodal framework that harnesses the interaction between textual and visual information for the analysis of social media content. The key innovations of our work include the application of LLM-based summarization for refined text preprocessing, the use of domain-aware image captioning to extract relevant visual features, and the implementation of a specialized joint modeling approach that captures cross-modal relationships. Another critical aspect of our framework is its parameter-efficient design. By optimizing the architecture to balance performance and computational cost, the model remains feasible for deployment in real-world scenarios where resources may be limited. This practical consideration makes the framework suitable for integration into social media monitoring systems and fact-checking platforms that require both accuracy and efficiency. Overall, the results underscore the importance of moving beyond text-only models and embracing multimodal strategies to accurately interpret the complex and diverse nature of user-generated content in social media environments.
\bibliographystyle{elsarticle-num}
\bibliography{biblography}

\begin{thebibliography}{10}
\expandafter\ifx\csname url\endcsname\relax
  \def\url#1{\texttt{#1}}\fi
\expandafter\ifx\csname urlprefix\endcsname\relax\def\urlprefix{URL }\fi
\expandafter\ifx\csname href\endcsname\relax
  \def\href#1#2{#2} \def\path#1{#1}\fi

\bibitem{PANGTEY2025111109}
L.~Pangtey, M.~Z.~U. Rehman, P.~Chaudhari, S.~Bansal, N.~Kumar, Emotion-aware dual cross-attentive neural network with label fusion for stance detection in misinformative social media content, Engineering Applications of Artificial Intelligence 156 (2025) 111109.
\newblock \href {http://dx.doi.org/https://doi.org/10.1016/j.engappai.2025.111109} {\path{doi:https://doi.org/10.1016/j.engappai.2025.111109}}.

\bibitem{Upadhyaya_Fisichella_Nejdl_2023}
A.~Upadhyaya, M.~Fisichella, W.~Nejdl, A multi-task model for sentiment aided stance detection of climate change tweets, Proceedings of the International AAAI Conference on Web and Social Media 17~(1) (2023) 854--865.
\newblock \href {http://dx.doi.org/10.1609/icwsm.v17i1.22194} {\path{doi:10.1609/icwsm.v17i1.22194}}.

\bibitem{10.1145/3543873.3587643}
N.~Salek~Faramarzi, F.~Hashemi~Chaleshtori, H.~Shirazi, I.~Ray, R.~Banerjee, Claim extraction and dynamic stance detection in covid-19 tweets, in: Companion Proceedings of the ACM Web Conference 2023, WWW '23 Companion, Association for Computing Machinery, New York, NY, USA, 2023, p. 1059–1068.
\newblock \href {http://dx.doi.org/10.1145/3543873.3587643} {\path{doi:10.1145/3543873.3587643}}.

\bibitem{MARTINEZ2023103294}
R.~Y. Martínez, G.~Blanco, A.~Lourenço, Spanish corpora of tweets about covid-19 vaccination for automatic stance detection, Information Processing \& Management 60~(3) (2023) 103294.
\newblock \href {http://dx.doi.org/https://doi.org/10.1016/j.ipm.2023.103294} {\path{doi:https://doi.org/10.1016/j.ipm.2023.103294}}.

\bibitem{rehman2025implihatevid}
M.~Z.~U. Rehman, A.~Bhatnagar, O.~Kabde, S.~Bansal, N.~Kumar, Implihatevid: A benchmark dataset and two-stage contrastive learning framework for implicit hate speech detection in videos, in: Proceedings of the 63rd Annual Meeting of the Association for Computational Linguistics (Volume 1: Long Papers), 2025, pp. 17209--17221.

\bibitem{10.1145/3533430}
Y.~Zhang, D.~Ma, P.~Tiwari, C.~Zhang, M.~Masud, M.~Shorfuzzaman, D.~Song, Stance-level sarcasm detection with bert and stance-centered graph attention networks, ACM Trans. Internet Technol. 23~(2).

\bibitem{dadas-etal-2020-evaluation}
S.~Dadas, M.~Pere{\l}kiewicz, R.~Po{\'s}wiata, \href{https://aclanthology.org/2020.lrec-1.207/}{Evaluation of sentence representations in {P}olish}, in: N.~Calzolari, F.~B{\'e}chet, P.~Blache, K.~Choukri, C.~Cieri, T.~Declerck, S.~Goggi, H.~Isahara, B.~Maegaard, J.~Mariani, H.~Mazo, A.~Moreno, J.~Odijk, S.~Piperidis (Eds.), Proceedings of the Twelfth Language Resources and Evaluation Conference, European Language Resources Association, Marseille, France, 2020, pp. 1674--1680.
\newline\urlprefix\url{https://aclanthology.org/2020.lrec-1.207/}

\bibitem{10651178}
H.~Chen, K.~Yan, M.~R. Kadhim, K.~Wu, L.~Tian, Sentkb-bert: Sentiment-filtered knowledge-based stance detection, in: 2024 International Joint Conference on Neural Networks (IJCNN), 2024, pp. 1--7.
\newblock \href {http://dx.doi.org/10.1109/IJCNN60899.2024.10651178} {\path{doi:10.1109/IJCNN60899.2024.10651178}}.

\bibitem{CHENG2025113399}
Y.~Cheng, Q.~Zhang, C.~Shi, L.~Xiao, S.~Hao, L.~Hu, Cosd: Collaborative stance detection with contrastive heterogeneous topic graph learning, Knowledge-Based Systems 317 (2025) 113399.
\newblock \href {http://dx.doi.org/https://doi.org/10.1016/j.knosys.2025.113399} {\path{doi:https://doi.org/10.1016/j.knosys.2025.113399}}.

\bibitem{10654680}
M.~Yan, T.~Z. Joey, W.~T. Ivor, Collaborative knowledge infusion for low-resource stance detection, Big Data Mining and Analytics 7~(3) (2024) 682--698.
\newblock \href {http://dx.doi.org/10.26599/BDMA.2024.9020021} {\path{doi:10.26599/BDMA.2024.9020021}}.

\bibitem{DANGI2025110895}
S.~Dangi, S.~K. Mullapudi, C.~S. Raghaw, S.~S. Dar, M.~Z.~U. Rehman, N.~Kumar, \href{https://www.sciencedirect.com/science/article/pii/S0168169925010014}{A multi-temporal multi-spectral attention-augmented deep convolution neural network with contrastive learning for crop yield prediction}, Computers and Electronics in Agriculture 239 (2025) 110895.
\newblock \href {http://dx.doi.org/https://doi.org/10.1016/j.compag.2025.110895} {\path{doi:https://doi.org/10.1016/j.compag.2025.110895}}.
\newline\urlprefix\url{https://www.sciencedirect.com/science/article/pii/S0168169925010014}

\bibitem{10.1145/3589335.3651467}
K.-H. Kuo, M.-H. Wang, H.-Y. Kao, Y.-C. Dai, Advancing stance detection of political fan pages: A multimodal approach, in: Companion Proceedings of the ACM Web Conference 2024, WWW '24, Association for Computing Machinery, New York, NY, USA, 2024, p. 702–705.
\newblock \href {http://dx.doi.org/10.1145/3589335.3651467} {\path{doi:10.1145/3589335.3651467}}.

\bibitem{10120685}
S.~Sengan, S.~Vairavasundaram, L.~Ravi, A.~Q.~M. AlHamad, H.~A. Alkhazaleh, M.~Alharbi, Fake news detection using stance extracted multimodal fusion-based hybrid neural network, IEEE Transactions on Computational Social Systems 11~(4) (2024) 5146--5157.
\newblock \href {http://dx.doi.org/10.1109/TCSS.2023.3269087} {\path{doi:10.1109/TCSS.2023.3269087}}.

\bibitem{10098547}
P.~J. Khiabani, A.~Zubiaga, Few-shot learning for cross-target stance detection by aggregating multimodal embeddings, IEEE Transactions on Computational Social Systems 11~(2) (2024) 2081--2090.
\newblock \href {http://dx.doi.org/10.1109/TCSS.2023.3264114} {\path{doi:10.1109/TCSS.2023.3264114}}.

\bibitem{Lan_Gao_Jin_Li_2024}
X.~Lan, C.~Gao, D.~Jin, Y.~Li, \href{https://ojs.aaai.org/index.php/ICWSM/article/view/31360}{Stance detection with collaborative role-infused llm-based agents}, Proceedings of the International AAAI Conference on Web and Social Media 18~(1) (2024) 891--903.
\newblock \href {http://dx.doi.org/10.1609/icwsm.v18i1.31360} {\path{doi:10.1609/icwsm.v18i1.31360}}.
\newline\urlprefix\url{https://ojs.aaai.org/index.php/ICWSM/article/view/31360}

\bibitem{zhang-etal-2024-llm-driven}
Z.~Zhang, Y.~Li, J.~Zhang, H.~Xu, {LLM}-driven knowledge injection advances zero-shot and cross-target stance detection, in: K.~Duh, H.~Gomez, S.~Bethard (Eds.), Proceedings of the 2024 Conference of the North American Chapter of the Association for Computational Linguistics: Human Language Technologies (Volume 2: Short Papers), Association for Computational Linguistics, Mexico City, Mexico, 2024, pp. 371--378.
\newblock \href {http://dx.doi.org/10.18653/v1/2024.naacl-short.32} {\path{doi:10.18653/v1/2024.naacl-short.32}}.

\bibitem{10.1145/3716856}
R.~Yang, J.~Ma, W.~Gao, H.~Lin, Llm-enhanced multiple instance learning for joint rumor and stance detection with social context information, ACM Trans. Intell. Syst. Technol. 16~(3).
\newblock \href {http://dx.doi.org/10.1145/3716856} {\path{doi:10.1145/3716856}}.

\bibitem{10.1007/s10994-024-06587-y}
M.~Gambini, C.~Senette, T.~Fagni, M.~Tesconi, Evaluating large language models for user stance detection on x (twitter), Mach. Learn. 113~(10) (2024) 7243–7266.
\newblock \href {http://dx.doi.org/10.1007/s10994-024-06587-y} {\path{doi:10.1007/s10994-024-06587-y}}.

\bibitem{app15115809}
X.~Chen, B.~Liu, H.~Hu, Y.~Cai, M.~Guo, X.~Ma, Integrating graph neural networks and large language models for stance detection via heterogeneous stance networks, Applied Sciences 15~(11).
\newblock \href {http://dx.doi.org/10.3390/app15115809} {\path{doi:10.3390/app15115809}}.

\bibitem{11006205}
M.~Shafiei, H.~Rahmani, A.~Derakhshan, M.~Allahgholi, Caskow: Context-aware stance detection using external knowledge-augmented llm, in: 2025 11th International Conference on Web Research (ICWR), 2025, pp. 123--129.
\newblock \href {http://dx.doi.org/10.1109/ICWR65219.2025.11006205} {\path{doi:10.1109/ICWR65219.2025.11006205}}.

\bibitem{devlin-etal-2019-bert}
J.~Devlin, M.-W. Chang, K.~Lee, K.~Toutanova, {BERT}: Pre-training of deep bidirectional transformers for language understanding, in: J.~Burstein, C.~Doran, T.~Solorio (Eds.), Proceedings of the 2019 Conference of the North {A}merican Chapter of the Association for Computational Linguistics: Human Language Technologies, Volume 1 (Long and Short Papers), Association for Computational Linguistics, Minneapolis, Minnesota, 2019, pp. 4171--4186.
\newblock \href {http://dx.doi.org/10.18653/v1/N19-1423} {\path{doi:10.18653/v1/N19-1423}}.

\bibitem{DAR2025125337}
S.~S. Dar, M.~Z.~U. Rehman, K.~Bais, M.~A. Haseeb, N.~Kumar, A social context-aware graph-based multimodal attentive learning framework for disaster content classification during emergencies, Expert Systems with Applications 259 (2025) 125337.
\newblock \href {http://dx.doi.org/https://doi.org/10.1016/j.eswa.2024.125337} {\path{doi:https://doi.org/10.1016/j.eswa.2024.125337}}.

\bibitem{Bansal2025}
S.~Bansal, M.~Kumar, C.~S. Raghaw, N.~Kumar, Sentiment and hashtag-aware attentive deep neural network for multimodal post popularity prediction, Neural Computing and Applications 37~(4) (2025) 2799--2824.
\newblock \href {http://dx.doi.org/10.1007/s00521-024-10755-5} {\path{doi:10.1007/s00521-024-10755-5}}.

\bibitem{multimodal}
A.~Anshul, G.~S. Pranav, M.~Z.~U. Rehman, N.~Kumar, A multimodal framework for depression detection during covid-19 via harvesting social media, IEEE Transactions on Computational Social Systems 11~(2) (2024) 2872--2888.
\newblock \href {http://dx.doi.org/10.1109/TCSS.2023.3309229} {\path{doi:10.1109/TCSS.2023.3309229}}.

\bibitem{rehman2025multimodal}
M.~Z.~U. Rehman, D.~Raghuvanshi, U.~Jain, S.~Bansal, N.~Kumar, A multimodal-multitask framework with cross-modal relation and hierarchical interactive attention for semantic comprehension, Information Fusion (2025) 103628.\href {http://dx.doi.org/https://doi.org/10.1016/j.inffus.2025.103628} {\path{doi:https://doi.org/10.1016/j.inffus.2025.103628}}.

\bibitem{dar2025explainable}
S.~S. Dar, B.~Kaurav, A.~Jain, C.~S. Raghaw, M.~Z.~U. Rehman, N.~Kumar, An explainable deep neural network with frequency-aware channel and spatial refinement for flood prediction in sustainable cities, Sustainable Cities and Society (2025) 106480.

\bibitem{10.1145/3618057}
P.~Chaudhari, P.~Nandeshwar, S.~Bansal, N.~Kumar, Mahaemosen: Towards emotion-aware multimodal marathi sentiment analysis, ACM Trans. Asian Low-Resour. Lang. Inf. Process. 22~(9).
\newblock \href {http://dx.doi.org/10.1145/3618057} {\path{doi:10.1145/3618057}}.

\bibitem{10887864}
D.~Raghuvanshi, X.~Gao, Z.~Li, S.~Bansal, M.~Coler, N.~Kumar, S.~Nayak, Intra-modal relation and emotional incongruity learning using graph attention networks for multimodal sarcasm detection, in: ICASSP 2025 - 2025 IEEE International Conference on Acoustics, Speech and Signal Processing (ICASSP), 2025, pp. 1--5.
\newblock \href {http://dx.doi.org/10.1109/ICASSP49660.2025.10887864} {\path{doi:10.1109/ICASSP49660.2025.10887864}}.

\bibitem{DAR2024112526}
S.~S. Dar, M.~K. Karandikar, M.~Z.~U. Rehman, S.~Bansal, N.~Kumar, A contrastive topic-aware attentive framework with label encodings for post-disaster resource classification, Knowledge-Based Systems 304 (2024) 112526.
\newblock \href {http://dx.doi.org/https://doi.org/10.1016/j.knosys.2024.112526} {\path{doi:https://doi.org/10.1016/j.knosys.2024.112526}}.

\bibitem{wang-etal-2024-multiclimate}
J.~Wang, L.~Zuo, S.~Peng, B.~Plank, {M}ulti{C}limate: Multimodal stance detection on climate change videos, in: D.~Dementieva, O.~Ignat, Z.~Jin, R.~Mihalcea, G.~Piatti, J.~Tetreault, S.~Wilson, J.~Zhao (Eds.), Proceedings of the Third Workshop on NLP for Positive Impact, Association for Computational Linguistics, Miami, Florida, USA, 2024, pp. 315--326.
\newblock \href {http://dx.doi.org/10.18653/v1/2024.nlp4pi-1.27} {\path{doi:10.18653/v1/2024.nlp4pi-1.27}}.

\bibitem{resnet}
K.~He, X.~Zhang, S.~Ren, J.~Sun, Deep residual learning for image recognition, in: 2016 IEEE Conference on Computer Vision and Pattern Recognition (CVPR), 2016, pp. 770--778.
\newblock \href {http://dx.doi.org/10.1109/CVPR.2016.90} {\path{doi:10.1109/CVPR.2016.90}}.

\bibitem{vit}
A.~Dosovitskiy, L.~Beyer, A.~Kolesnikov, D.~Weissenborn, X.~Zhai, T.~Unterthiner, M.~Dehghani, M.~Minderer, G.~Heigold, S.~Gelly, J.~Uszkoreit, N.~Houlsby, \href{https://openreview.net/forum?id=YicbFdNTTy}{An image is worth 16x16 words: Transformers for image recognition at scale}, in: International Conference on Learning Representations, 2021.
\newline\urlprefix\url{https://openreview.net/forum?id=YicbFdNTTy}

\bibitem{blip}
J.~Li, D.~Li, C.~Xiong, S.~Hoi, \href{https://proceedings.mlr.press/v162/li22n.html}{{BLIP}: Bootstrapping language-image pre-training for unified vision-language understanding and generation}, in: K.~Chaudhuri, S.~Jegelka, L.~Song, C.~Szepesvari, G.~Niu, S.~Sabato (Eds.), Proceedings of the 39th International Conference on Machine Learning, Vol. 162 of Proceedings of Machine Learning Research, PMLR, 2022, pp. 12888--12900.
\newline\urlprefix\url{https://proceedings.mlr.press/v162/li22n.html}

\bibitem{clip}
A.~Radford, J.~W. Kim, C.~Hallacy, A.~Ramesh, G.~Goh, S.~Agarwal, G.~Sastry, A.~Askell, P.~Mishkin, J.~Clark, G.~Krueger, I.~Sutskever, \href{https://proceedings.mlr.press/v139/radford21a.html}{Learning transferable visual models from natural language supervision}, in: M.~Meila, T.~Zhang (Eds.), Proceedings of the 38th International Conference on Machine Learning, Vol. 139 of Proceedings of Machine Learning Research, PMLR, 2021, pp. 8748--8763.
\newline\urlprefix\url{https://proceedings.mlr.press/v139/radford21a.html}

\bibitem{fakefind}
S.~Sengan, S.~Vairavasundaram, L.~Ravi, A.~Q.~M. AlHamad, H.~A. Alkhazaleh, M.~Alharbi, Fake news detection using stance extracted multimodal fusion-based hybrid neural network, IEEE Transactions on Computational Social Systems 11~(4) (2024) 5146--5157.
\newblock \href {http://dx.doi.org/10.1109/TCSS.2023.3269087} {\path{doi:10.1109/TCSS.2023.3269087}}.

\bibitem{shah-etal-2024-memeclip}
S.~B. Shah, S.~Shiwakoti, M.~Chaudhary, H.~Wang, {M}eme{CLIP}: Leveraging {CLIP} representations for multimodal meme classification, in: Y.~Al-Onaizan, M.~Bansal, Y.-N. Chen (Eds.), Proceedings of the 2024 Conference on Empirical Methods in Natural Language Processing, Association for Computational Linguistics, Miami, Florida, USA, 2024, pp. 17320--17332.
\newblock \href {http://dx.doi.org/10.18653/v1/2024.emnlp-main.959} {\path{doi:10.18653/v1/2024.emnlp-main.959}}.

\bibitem{REHMAN2025103895}
M.~Z.~U. Rehman, S.~Zahoor, A.~Manzoor, M.~Maqbool, N.~Kumar, A context-aware attention and graph neural network-based multimodal framework for misogyny detection, Information Processing \& Management 62~(1) (2025) 103895.
\newblock \href {http://dx.doi.org/https://doi.org/10.1016/j.ipm.2024.103895} {\path{doi:https://doi.org/10.1016/j.ipm.2024.103895}}.

\bibitem{RAGHAW2025107363}
C.~S. Raghaw, A.~Yadav, J.~S. Sanjotra, S.~Dangi, N.~Kumar, Mnet-sat: A multiscale network with spatial-enhanced attention for segmentation of polyps in colonoscopy, Biomedical Signal Processing and Control 102 (2025) 107363.
\newblock \href {http://dx.doi.org/https://doi.org/10.1016/j.bspc.2024.107363} {\path{doi:https://doi.org/10.1016/j.bspc.2024.107363}}.

\bibitem{bansal2025retrieval}
S.~Bansal, S.~Parimala, N.~Kumar, Retrieval augmented encoder-decoder with diffusion for sequential hashtag recommendation in disaster events, in: Proceedings of the International AAAI Conference on Web and Social Media, Vol.~19, 2025, pp. 206--225.

\bibitem{RAGHAW2025108225}
C.~S. Raghaw, J.~S. Sanjotra, M.~Z.~U. Rehman, S.~Bansal, S.~S. Dar, N.~Kumar, \href{https://www.sciencedirect.com/science/article/pii/S1746809425007360}{T-mpednet: Unveiling the synergy of transformer-aware multiscale progressive encoder-decoder network with feature recalibration for tumor and liver segmentation}, Biomedical Signal Processing and Control 110 (2025) 108225.
\newblock \href {http://dx.doi.org/https://doi.org/10.1016/j.bspc.2025.108225} {\path{doi:https://doi.org/10.1016/j.bspc.2025.108225}}.
\newline\urlprefix\url{https://www.sciencedirect.com/science/article/pii/S1746809425007360}

\end{thebibliography}
\end{document}